\documentclass[conference]{IEEEtran}

\usepackage{graphicx} 
\usepackage{subfigure}
\usepackage{amssymb}
\usepackage{amsfonts}
\usepackage{amsmath}
\usepackage{amsthm}
\usepackage{latexsym}
\usepackage{url}
\usepackage{enumerate}
\usepackage{wrapfig}

\usepackage{algorithm}
\usepackage{algorithmic}

\usepackage{tabularx}
\usepackage{array}

\usepackage{picinpar}
\usepackage{pgf}
\DeclareMathOperator*{\argmin}{\mathrm{argmin}}

\newcommand{\supp}{{\mathrm{supp}}}

\def \bD{{\mathbf{D}}}
\def \bH{{\mathbf{H}}}
\def \bX{{\mathbf{X}}}
\def \bZ{{\mathbf{Z}}}

\def \bd{{\mathbf{d}}}

\def \br{{\mathbf{r}}}

\def \bx{{\mathbf{x}}}

\def \bz{{\mathbf{z}}}

\def \supp{{\mathrm{supp}}}

\begin{document}
%


\title{Stochastic Coordinate Coding and Its Application for Drosophila Gene Expression Pattern Annotation}
%
%
%
%
%

%
\author{
%
%
Binbin Lin$^1$, Qingyang Li$^1$ , Qian Sun$^1$ , Ming-Jun Lai$^2$ , Ian Davidson$^3$ , Wei Fan$^4$ , Jieping Ye$^1$ \\
       {$^1$Center for Evolutionary Medicine and Informatics, The Biodesign Institute, ASU, Tempe, AZ}\\
     {$^2$Department of Mathematics, University of Georgia, Athens, GA}\\
      {$^3$Department of Computer Science, University of California, Davis, CA}\\
     {$^4$Noah's Ark Lab, Huawei Technologies Co. Ltd., Sha Tin, Hong Kong}
}


\maketitle
\begin{abstract}
\emph{Drosophila melanogaster} has been established as a model organism for investigating the fundamental principles of developmental gene interactions. The gene expression patterns of {Drosophila melanogaster} can be documented as digital images, which are annotated with anatomical ontology terms to facilitate pattern discovery and comparison. The automated annotation of gene expression pattern images has received increasing attention due to the recent expansion of the image database. The effectiveness of gene expression pattern annotation relies on the quality of feature representation. Previous studies have demonstrated that sparse coding is effective for extracting features from gene expression images. However, solving sparse coding remains a computationally challenging problem, especially when dealing with large-scale data sets and learning large size dictionaries. In this paper, we propose a novel algorithm to solve the sparse coding problem, called Stochastic Coordinate Coding (SCC). The proposed algorithm alternatively updates the sparse codes via just a few steps of coordinate descent and updates the dictionary via second order stochastic gradient descent. The computational cost is further reduced by focusing on the non-zero components of the sparse codes and the corresponding columns of the dictionary only in the updating procedure. Thus, the proposed algorithm significantly improves the efficiency and the scalability, making sparse coding applicable for large-scale data sets and large dictionary sizes. Our experiments on Drosophila gene expression data sets demonstrate the efficiency and the effectiveness of the proposed algorithm.
\end{abstract}


\section{Introduction}
\emph{Drosophila melanogaster} has been established as a model organism for
investigating the fundamental principles of developmental gene
interactions~\cite{SeanCarroll2005, Levine2005, Matthews2005}. The Berkeley
Drosophila Genome Project (BDGP,~\cite{Tomancak02Sys, Tomancak:global:full}) has
produced a comprehensive atlas of gene expression patterns in the form of
digital images by RNA \emph{in situ} hybridization~\cite{GaryGrumbling:FlyBase}
in order to facilitate a deep understanding of transcriptional regulation during
Drosophila embryogenesis. The images in BDGP are annotated with anatomical and
developmental ontology terms using a controlled vocabulary~\cite{Tomancak02Sys}.
To facilitate pattern discovery and comparison, many web-based resources have
been created to conduct comparative analysis based on the body part keywords and
the associated images. Currently, the annotation is performed manually by human
experts. With the increasing size of available images generated by high
throughput technologies, it is imperative to design efficient and effective computational
methods to automatically annotate images capturing spatial patterns of gene
expression.

\begin{figure}\centering
\includegraphics[width=0.95\linewidth]{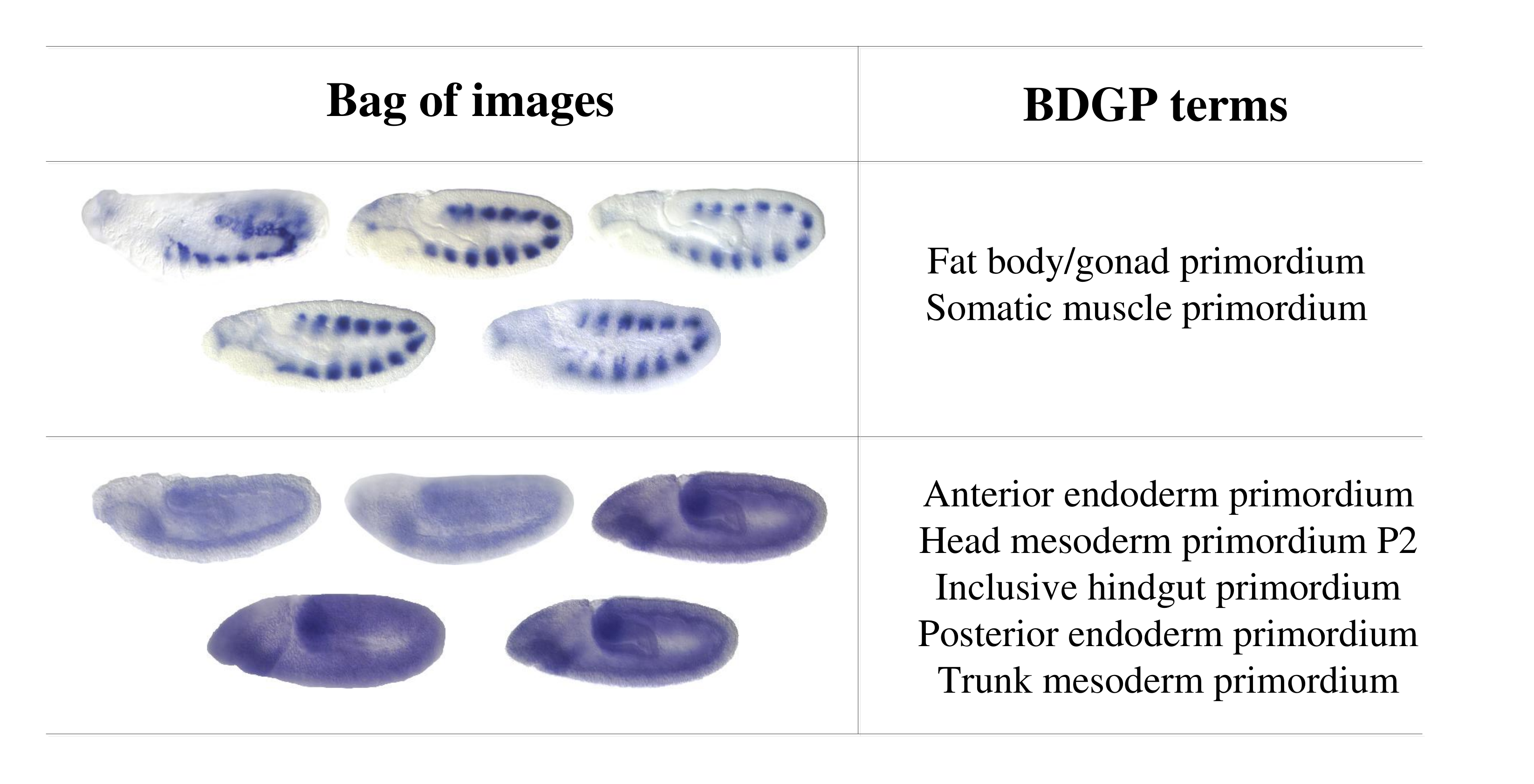}
\caption{Sample bag (groups) of images and the associated terms in BDGP database.}\label{fig:in}
\end{figure}

The BDGP gene expression pattern annotation problem can be formulated as an
image annotation problem, which has been widely studied in computer vision and
machine learning. In particular, a collection of images from the same
developmental stage range and the same gene are annotated by a sub-set of the
keywords (see Fig.~\ref{fig:in}). Although traditional image annotation
methodologies can be employed to solve this problem, significant challenges
remain due to the multi-instance multi-label nature of this problem. Since the
annotation associated with a group of images does not imply an association with all the
images in this group, we need to develop approaches to retain the group
membership information. Due to the effects of stochastic processes during
embryogenesis, no two embryos develop identically. And the current image
acquisition techniques limit the quality of the images. Thus, the shape and
color of the same body part may vary from image to image. Invariance to local
distortions is required for an accurate annotation system. Several prior works
on the automatic annotation of \emph{Drosophila} gene expression images have
been reported. Zhou and Peng~\cite{JieZhou} constructed their system based on
the assumption that each image in the group is annotated by all the terms
assigned to that group; Ji et al.~\cite{Ji:KDD09a} considered a learning
framework that incorporates the term-term interactions; Yuan et
al.~\cite{yuan2012learning} and Sun et al.~\cite{sun2013image} adopted sparse
coding for image annotation with the dictionary generated by the
clustering techniques. However, the dictionary is fixed during the training process
due to the expensive learning cost. This motivates us to develop an efficient sparse coding algorithm to efficiently learn the dictionary and sparse
feature representations from the data.


Sparse coding concerns the problem of reconstructing data vectors using sparse linear combinations of basis vectors~\cite{olshausen1996emergence,chen1998atomic,donoho2003optimally}. It has become extremely popular for learning the dictionary and extracting features from images in the last decade. Sparse coding has been applied in many fields including audio processing \cite{Smith2006}, text mining \cite{BalakrishnanM08} and image recognition \cite{SzlamGL12}. Different from traditional feature extraction methods like principal component analysis and its variants, sparse coding learns non-orthogonal and over-complete dictionaries which have more flexibility to represent the data. Sparse coding can also model inhibition between the bases by sparsifying their activations. Similar properties have been observed in biological neurons, thus making sparse coding a plausible model of the visual cortex~\cite{olshausen1997sparse,olshausen2004sparse}.

Despite the rich promise of sparse coding models, sparse coding is computationally expensive especially when dealing with large-scale data. The main computational cost of sparse coding lies in the updating of sparse codes and the dictionary. It is known that updating the sparse code is usually much more time consuming. Therefore, much of recent work has been devoted to seeking efficient optimization algorithms for updating the sparse code~\cite{lee2006efficient, tibshirani2012strong}. The basic idea of these methods is to quickly identify the non-zero entries of the sparse code, thus reducing the search space. However, most of these algorithms are iterative batch methods which may not scale to very large data sets~\cite{bottou201113}, since updating the dictionary involves the computation of the full gradient of the dictionary from the whole data set and is expensive. Recently, several work based on stochastic gradient descent and online learning has been proposed. \cite{mairal2009online} proposed an online dictionary learning algorithm which updates the dictionary for each incoming data point. It is expected that the dictionary will converge faster in the online setting. However, even when the dictionary has been learned, one has to further learn the sparse code, which is also computationally expensive especially for large-scale data sets.

In this paper, we propose a novel approach for efficiently solving the sparse coding problem.
The key ingredients of the proposed SCC algorithm involve:
\begin{enumerate}
\item When updating the sparse code, we only perform a few steps of coordinate descent. For each image patch, we further speed up the coordinate descent by updating only the support of the sparse code.
\item When updating the dictionary, we only update the columns corresponding to the support of the sparse code.
\item When doing stochastic gradient descent, we choose an adaptive learning rate which speeds up the convergence of the algorithm.
\end{enumerate}
Extensive experiments on Drosophila gene expression data sets demonstrate the efficiency of the proposed algorithm. 

\section{Backgrounds and Related Work}
In this section, we review sparse coding and related work.

We first introduce our notation used throughout this paper. We use boldface lower case letters (e.g., $\bx, \bz$) to denote vectors and use blodface upper case letters (e.g., $\bX, \bZ$) to denote matrices. Scalars are denoted by lower or upper case letters (e.g., $p, M$). Given a data set $\bX = ( \bx_1 \cdots \bx_n )$ of image patches, each image patch is a $p$-dimensional vector, i.e., $\bx_i \in \mathbb{R}^p $, $i=1, \ldots, n$. Moreover, each $\bx_i$ is preprocessed to be zero mean and unit $l_2$ norm. We first extract meaningful features from these image patches using sparse coding. The learned features will be used for image annotation.

The linear decomposition of an image patch using a few number of basis or atoms of a learned dictionary has recently led to state-of-art performance in numerous signal processing and machine learning tasks. Specifically, suppose there are $m$ atoms $\bd_j\in \mathbb{R}^p, j=1, \ldots, m$, where the number of atoms is usually much smaller than the number of image patches $n$ but larger than the dimension of the image patch $p$. Each image patch can then be represented as
$
\bx_i = \sum_{j=1}^m z_{i,j} \bd_j.
$
Therefore, each $p$-dimensional image patch $\bx_i$ is represented by a $m$-dimensional vector
$\bz_i = (z_{i,1}, \ldots, z_{i,m})^T$. It is further assumed that each image patch can be represented only by a small group of atoms, that is, the learned feature vector $\bz_i$ is a sparse vector.

Given one image patch $\bx_i$, one can formularize the above idea as the following optimization problem:
\begin{equation}\label{eq:sc-obj-one}
\min f_i(\bD, \bz_i) = \frac{1}{2}\| \bD \bz_i - \bx_i \|^2 + \lambda \| \bz_i \|_1,
\end{equation}
where $\lambda$ is the regularization parameter, $\| \cdot \|$ is the standard Euclidean norm and $\| \bz_i \|_1 = \sum_{j=1}^m |z_{i,j}|$.
The first term of Eq.\eqref{eq:sc-obj-one} is the reconstruction error, which measures how well the new feature represents the image patch. The second term of Eq.\eqref{eq:sc-obj-one} ensures the sparsity of the learned feature $\bz_i$. Each $\bz_i$ is often called the \emph{sparse code}. Since $\bz_i$ is sparse, there are only a few entries in $\bz_i$ which are non-zero. We call its non-zero entries as its \emph{support}, i.e., $\supp(\bz_i) = \{ z_{i,j}: z_{i,j} \neq 0, j=1,\cdots,m. \}$. Here $\bD = (\bd_1 \cdots \bd_m) \in \mathbb{R}^{m \times p} $ is called the \emph{dictionary}. To prevent an arbitrary scaling of the sparse code, each column of $\bD$ is restricted to be in a unit ball, i.e., $\| \bd_j \| \leq 1$. Given the whole data set $\bX = ( \bx_1 \cdots \bx_n )$, the sparse coding problem is then given as follows:
\begin{equation}\label{eq:sc-obj-all}
\min_{ \bD \in B_m, \bz_1, \cdots, \bz_m} \mathcal{F}( \bD, \bz_1, \cdots, \bz_m) \equiv \frac{1}{n}\sum_{i=1}^n f_i( \bD , \bz_i),
\end{equation}
where $B_m$ is the feasible set of $\bD$ which is defined as follows:
\[
B_m = \{ \bD\in \mathbb{R}^{p\times m}:  \forall j=1,\ldots, m, \| \bd_j \|_2 \leq 1 \}.
\]

It is a non-convex problem with respect to joint parameters in the dictionary $\bD$ and the sparse codes $\bZ = (\bz_1 \cdots \bz_n)$. Therefore, it is often difficult to find a global optimum. However, it is a convex problem when either $\bD$ or $\bZ$ is fixed. When the dictionary $\bD$ is fixed, solving each sparse code $\bz_i$ is the well known lasso problem \cite{tibshirani96regression}. Many methods have been proposed to solve this problem, including Least Angle Regression (LARS, \cite{efron2004least}), Fast Iterative Soft-Thresholding Algorithm (FISTA, \cite{Beck:2009:FISTA}) and Coordinate Descent (CD, \cite{TTW08a}). It might be worth noting that when the feature dimension $m$ is large which is often the case, solving a lasso problem is very time consuming. When the sparse codes are fixed, it is a simple quadratic problem. Therefore, one often uses an alternating optimization approach to solve the sparse coding problem. Specifically, when $\bD$ is fixed, we update the sparse code $\bz_i$ for each image patch $\bx_i$. When the sparse codes are fixed, we use gradient descent to update the dictionary:
$$
\bD  \leftarrow \bD - \eta \frac{1}{n}\sum_{i=1}^n \nabla_{\bD} f_i(\bD, \bz_i) = \bD - \eta \frac{1}{n}\sum_{i=1}^n (\bD\bz_i - \bx_i) \bz_i^T,
$$
where $\eta$ is the step size. However, at each iteration, full gradient descent requires evaluation of $n$ derivatives, which is very expensive when the data set is of large-scale. A popular modification is Stochastic Gradient Descent (SGD, see~\cite{bottou1998online}). At each iteration, we randomly draw an image patch $\bx_t$, and update the dictionary as follows:
\[
\bD_{t+1} \leftarrow \bD_{t} - \eta_t \nabla_{\bD_{t}} f_t(\bD_{t}, \bz_{{t}}),
\]
where $ \eta_t$ is called the learning rate. 

We summarize the optimization methods in the following. First we initialize the dictionary $\bD$. Many dictionary initialization methods have been proposed, such as random weights~\cite{JarrettKRL09}, random patches and k-means. A detailed comparison of the performance among these initialization methods has been discussed in \cite{ICML2011Coates}. With the initial dictionary, conventional sparse coding algorithms include the following main steps:
\begin{enumerate}
\item Get an image patch $\bx_i$.
\item Calculate the sparse code $\bz_i$ by using LARS, FISTA or coordinate descent.
\item Update the dictionary $\bD$ by performing stochastic gradient descent. 
\item Go to step 1 and iterate.
\end{enumerate}
We call each cycle, i.e. each image patch has been trained once, as an \emph{epoch}. Usually, several epochs are required to obtain a satisfactory result. When the number of image patches and the dictionary size is large, step 2 and step 3 are still very slow. We propose a novel algorithm to improve both of these parts, which is presented in the next section.

\section{Stochastic Coordinate Coding}

In this section, we introduce our Stochastic Coordinate Coding (SCC) algorithm. It is known that solving the sparse coding problem usually is very time consuming especially when dealing with large-scale data sets and large size dictionaries~\cite{lee2006efficient}. The proposed algorithm aims to dramatically reduce the computational cost of the sparse coding while keeping comparable performance.

\begin{figure}\centering
\includegraphics[width=0.95\linewidth]{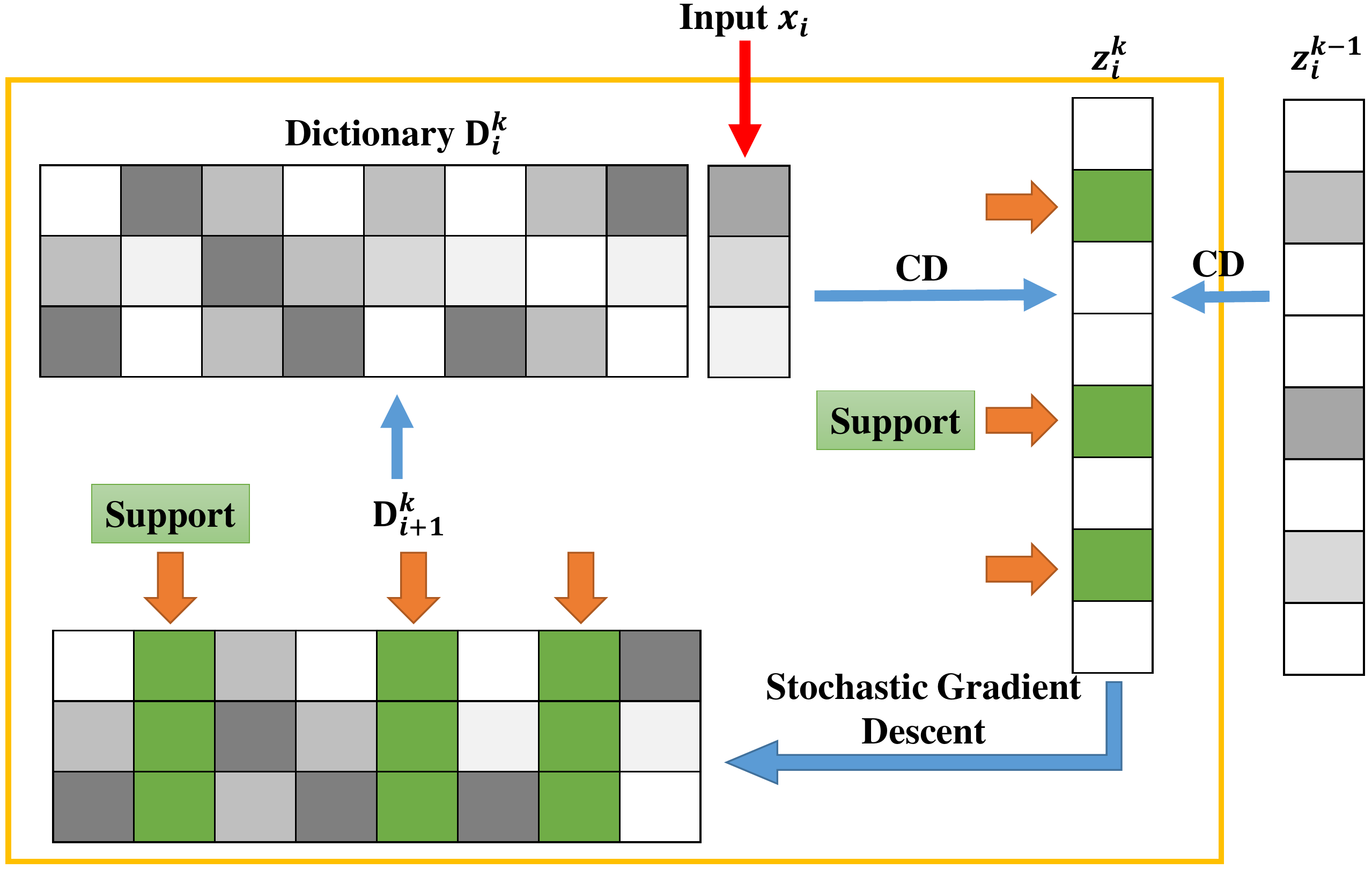}
\caption{Illustration of our algorithmic framework. With an image patch $\bx_i$, we perform one step of coordinate descent to find the support the sparse code. Next, we perform a few steps of coordinate descent on the support to obtain a new sparse code $\bz_i^{k}$. Then we update the support of the dictionary by second order stochastic gradient descent to obtain a new dictionary $\bD^k_{i+1}$.}\label{fig:alg-frame}
\end{figure}

We detail our algorithm in the following. Initialize the dictionary via any initialization method and denote it as $\bD_1^1$. Initialize the sparse code $\bz_i^{0} = 0$ for $i=1, \cdots, n$. Here we use superscript to represent the number of epochs and we use subscript to represent the index of data points. Then starting from $k=1$ and $i=1$, we do the following:
\begin{enumerate}
\item Get an image patch $\bx_i$
\item Update $\bz_i^k$ via one or a few steps of coordinate descent:
\begin{equation}\label{eq:cd-obj-a}
\bz_i^k = \mathrm{CD}( \bD^k_i, \bz_i^{k-1},\bx_i ).
\end{equation}
Specifically, for $j$ from 1 to $m$, we update the $j$-th coordinate $z_{i,j}^{k-1}$ of $\bz_i^{k-1}$ cyclicly as follows:
\[
\begin{aligned}
b_j &\leftarrow (\bd^k_{i,j})^T (\bx_i - \bD^k_i  \bz^{k-1}_i)+ z_{i,j}^{k-1}, \\
z_{i,j}^{k-1}  &\leftarrow  h_{\lambda}( b_j ),
\end{aligned}
\]
where $h$ is the soft thresholding shrinkage function~\cite{Combettes2005-MMS}. We call such one updating cycle as \emph{one step} of coordinate descent. The updated sparse code is then denoted by $\bz^{k}_i$. A detailed derivation of coordinate descent can be found in the appendix A.

%
\item Update the dictionary $\bD$ by using stochastic gradient descent:
\begin{equation}\label{eq:sgd}
\bD^k_{i+1} = P_{B_m}(\bD_{i}^k - \eta_{i}^k \nabla_{\bD_{i}^k} f_i(\bD_{i}^k, \bz_i^k)),
\end{equation}
where $P$ denotes the projection operator.  We set the learning rate as an approximation of the inverse of the Hessian matrix. The gradient of $D_{i}^k$ can be obtained as follows:
\[
\nabla_{\bD_{i}^k} f_i(\bD_{i}^k, \bz_i^k) = (D_{i}^k \bz^k_i - \bx_i) (\bz^k_i)^T.
\]
\item $i=i+1$. If $i>n$, then set $\bD^{k+1}_1 = \bD^k_{n+1}$, $k=k+1$ and $i=1$.
\end{enumerate}
We illustrate our algorithmic framework in Fig.~\eqref{fig:alg-frame}. At each iteration, we get an image patch $\bx_i$. Then we perform one or a few steps of coordinate descent to find the support of the sparse code. Next, we perform a few steps of coordinate descent on the support to obtain a new sparse code $\bz_i^{k}$. Then we update the support of the dictionary by second order stochastic gradient descent.

It is known that the second step - updating the sparse code is the most time consuming part~\cite{balasubramanian2013smooth}. Coordinate descent is known as one of the state of art methods for solving this lasso problem. Given an image patch $\bx_i$, coordinate descent initialize $\bz_i^0 = 0$ and then update the sparse code many times via matrix-vector multiplication and thresholding. Empirically, the iteration may take tens hundreds steps to converge. However, we observed that after a few steps, the support of the coordinates, i.e., the locations of the nonzero entries in $\bz_i$, is very accurate, usually less then ten steps. Note that the support of the sparse code is usually more important than the exact value of the sparse code. Moreover, since the original sparse coding is a non-convex problem and it involves an alternating updating, we do not need to run the coordinate descent to final convergence. Therefore, we propose to update the sparse code $\bz_i$ by using a few steps of coordinate descent. For the $k$-th epoch, we denote the updated sparse code as $\bz^k_i$. It will be used as an initial sparse code for the $k+1$-th epoch.

After updating the sparse code, we know its support. One of our key insights is that when updating the dictionary, we can only need to focus on the support of the dictionary but not all columns of the dictionary. Let $\bz_{i, j}^k$ denote $j$-th entry of $\bz_i^k$ and let $\bd_{i,j}^k$ denote the $j$-th column of the dictionary $\bD_{i}^k$. If $z_{i, j}^k = 0$, then $\nabla_{\bd_{i,j}^k} f_i(\bD_{i}^k, \bz_i^k) = (\bD_{i}^k \bz^k_i - \bx_i) z^k_{i,j} = 0$. Therefore, $\bd_{i,j}^k$ does not need to be updated. Assume $z_{i, j}^k$ is non-zero. Let $\bd_{i+1,j}^k$ denote the $j$-th column of the dictionary $\bD_{i+1}^k$. Then we can update $\bd_{i+1,j}^k$ as follows:
\begin{equation}\label{eq:update-dict}
\bd_{i+1, j}^k \leftarrow \bd_{i, j}^k -  \eta_{i,j}^k \nabla_{\bd_{i, j}^k} f_i(\bD^k_i, \bz_i^k) = \bd^k_{i,j} - \eta_{i,j}^k  z_{i,j}(\bD^k_i\bz_i^k - \bx_i),
\end{equation}
Note that $\bz_i^k$ here is a sparse vector, therefore computing $D^k_i\bz_i^k$ is very efficient. The computational cost will be significantly reduced when the support is very small. Note that for online dictionary learning, one usually has to update all columns of the dictionary. It is because that online dictionary learning uses the averaged gradient, which is usually not sparse. In other words, the support of the dictionary is itself. Therefore, one has to update all columns of the dictionary for each image patch. It is time consuming especially when the dictionary size is very large.

When the data sets are very large, the learning rate $\eta^k_i$ will be very small after going through large number of image patches. In this case, the dictionary will not change very much and the efficiency of the training will decrease. In practice, turning the learning rate is very tricky and sensitive. In this paper, we use an adaptive learning rate. We aim to design a learning rate with the following two principals. The first one is that for different columns of the dictionary, we may use different learning rates. The second is that for the same column, the learning rate should decrease. Otherwise, the algorithm might not converge. To obtain the learning rate, we use the Hessian matrix of the objective function. It can be shown that the following matrix provides an approximation of the Hessian:
$
\bH = \sum_{k,i} \bz^k_i(\bz^k_i)^T,
$
when $k$ and $i$ go to infinity. According to the second order stochastic gradient descent, we should use the inverse matrix of the Hessian as the learning rate. However, computing a matrix inversion problem is computationally expensive. In order to obtain the learning rate, we simply use the diagonal element of the matrix $\bH$. Note that if the columns of the dictionary have low correlation, $\bH$ is close to a diagonal matrix. Specifically, we first initialize $\bH = 0$. Then update the matrix $\bH$ as follows:
\begin{eqnarray}
\bH &\leftarrow& \bH + \bz^k_i(\bz_i^k)^T.
\end{eqnarray}
When updating the $j$-th column for the $i$th image patch $\bx_i$, we replace $\eta_{i, j}^k$ in Eq.~\eqref{eq:update-dict} by $1/h_{jj}$, where $h_{jj}$ is the $j$-th diagonal element of $\bH$.
In this way, we do not have to tune the learning rate parameter. It might be worth noting that we do not have to store the whole matrix of $\bH$ but only its diagonal elements. We summarize our algorithm in Algorithm \ref{alg:scc}.

\begin{algorithm}[ht!]
  \caption{SCC (Stochastic Coordinate Coding)}\label{alg:scc}
  \begin{algorithmic}
    \REQUIRE Data set $\bX=(\bx_1 \cdots \bx_n)\in\mathbb{R}^{p \times n}$
    \ENSURE $\bD \in \mathbb{R}^{p\times m}$ and $\bZ = (\bz_1 \cdots \bz_n) \in \mathbb{R}^{m\times n}$\\
    \textbf{Initialize:} $\bD_1^1$, $\bH = 0$ and $\bz_i^0 =0$ for $i=1, \ldots, n$.\\

   \textbf{for $k=1$ to $\kappa$ do} \\
        \quad  \textbf{for $i=1$ to $n$ do}\\
    \quad Get an image patch $\bx_i$ \\
    \quad Update $\bz_i^k$ via one or a few steps of coordinate descent:
    \[
    \bz_i^k \leftarrow \mathrm{CD}( \bD_{i}^k, \bz_i^{k-1}, \bx_i ).
    \]
    \quad Update the Hessian matrix and the learning rate:
    \[
    \bH \leftarrow \bH + \bz^k_i(\bz_i^k)^T, \quad \eta_{i, j}^k = 1/h_{jj}.
    \]
    \quad Update the support of the dictionary via SGD:
    \[
    \bd_{i+1, j}^k \leftarrow \bd^k_{i,j} - \eta_{i,j}^k  z_{i,j}(\bD^k_i\bz_i^k - \bx_i).
    \]
    \quad If $i=n$, set $\bD^{k+1}_1 = \bD^k_{n+1}$.\\
    \quad \textbf{end for} \\
    \textbf{end for} \\
    \textbf{Output} \\
    $\bD = \bD_n^\kappa$ and $\bz_i = \bz^\kappa_i$ for $i=1, \ldots, n$.
  \end{algorithmic}
\end{algorithm}

\section{Experiments}
In this section, we empirically evaluate the efficiency and effectiveness of our proposed Stochastic Coordinate Coding (SCC) algorithm. A detailed description of data and experimental setting is given in Section~\ref{sec:data-setting}. We study the influence of different settings of SCC in Section \ref{sec:model-selection}, including the influence of the number of coordinate descent steps and the learning rate. Finally, we compare SCC with the state-of-art sparse coding algorithm - Online dictionary Learning (OL, \cite{mairal2009online}) in terms of speed-up and accuracy in Section 5.3 and 5.4.

\subsection{Data Description and Experimental Setting}\label{sec:data-setting}
The Drosophila gene expression images used in our work are obtained from the FlyExpress database, which contains standardized images from the Berkeley Drosophila Genome Project (BDGP). The Drosophila embryogenesis is partitioned into 6 stage ranges (1-3, 4-6, 7-8, 9-10, 11-12, 13-17) in BDGP. We focus on the later 5 stage ranges as there are few keywords appeared in the first stage range.

The Drosophila embryos are 3D objects~\cite{weber2009visual}, and the FlyExpress database contains 2D images that are taken
from different views (lateral, dorsal, and lateral-dorsal)~\cite{mace2010extraction}. As majority of images in the database are in lateral view~\cite{ji2009drosophila}, we focus on the lateral-view images in our study. For each image, we first use a $16\times 16$ window to obtain a collection of small image patches. Then we extract a 128-dimensional Scale-Invariant Feature Transform (SIFT, \cite{lowe1999object}) feature from each image patch.  Each SIFT feature is further normalized to be zero mean and unit $l_2$ norm. The patches with small standard deviations were discarded. After preprocessing the data, we have 555009, 259882, 286349, 989653, 1006012 image patches for different stage ranges  (4-6, 7-8, 9-10, 11-12, 13-17) respectively.

For each state range, we first initialize the dictionary via selecting random patches~\cite{ICML2011Coates}, which has been shown to be a very efficient and effective initialization method in practice. Then we learn the sparse codes by different sparse coding methods using the same initial dictionary. All 5 stage ranges will be trained for 10 epochs using a batch size of 1. After learning the sparse codes, we apply max pooling~\cite{scherer2010evaluation} to generate the features for annotation. Finally, we employ the one-against-rest support vector machines (SVM, \cite{chang2011libsvm}) to annotate the gene expression pattern images.

The regularization parameter $\lambda$ is set to $0.10 = 1.2 / \sqrt{p}$ in all of our experiments. The $1/\sqrt{p}$ term comes from a classical normalization factor \cite{bickel2009simultaneous}, and the constant $1.2$ has shown to shown to produce about 10 non-zero coefficients for Drosophila data sets. We have implemented the proposed algorithm in C++ and all the experiments have been run on a single-CPU, eight-core 3.4Ghz machine.

\subsection{Model Selection}\label{sec:model-selection}
In this section, we study the influence of different algorithm settings, including the number of coordinate descent steps and learning rates.

\subsubsection{The Number of Coordinate Descent Steps}

\begin{figure}\centering
    \subfigure[Objective value curve]{\includegraphics[width=0.48\linewidth]{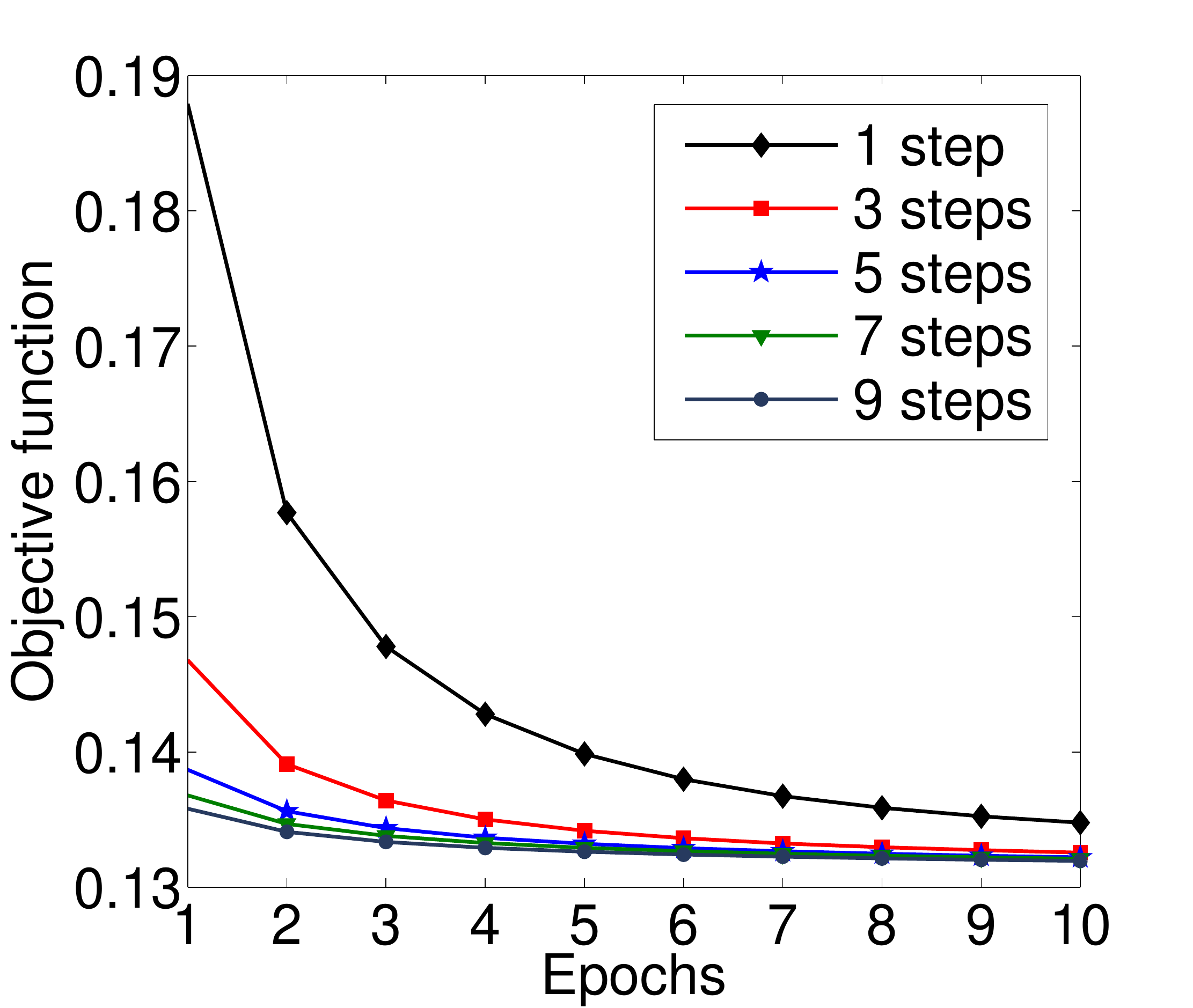}}
    \subfigure[Computational time]{\includegraphics[width=0.48\linewidth]{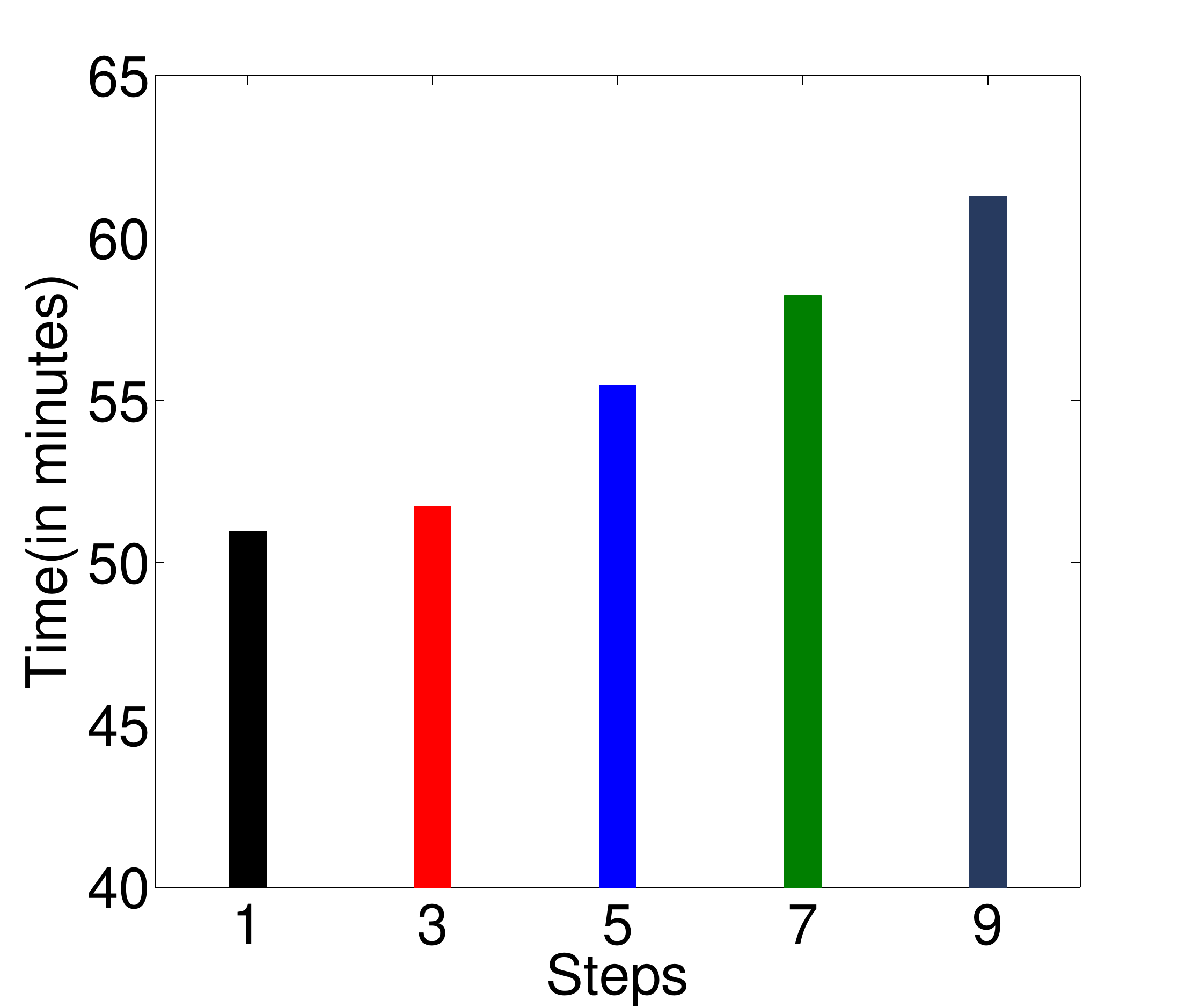}}
    \caption{{A comparison of different coordinate descent steps}. (a) shows the objective value curves when varying the number of coordinate descent steps. The horizontal axis represents the number of epochs. (b) shows the computational time (in minutes) of running 10 epochs. It can be seen from the figure that using a great number of coordinate descent steps can achieve lower objective value. However, the overall computational time would increase.}
    \label{fig:cd-steps}
\end{figure}

First, experiments were carried out to study the influence of the number of coordinate descent steps. We use the state range 2 in our experiments. It has 555009 image patches and each image patch is of 128 dimensions. The dictionary size is $1000\times 128$. We tested {1, 3, 5, 7, 9} steps of coordinate descent. The results are evaluated by the objective function value and the running time, as shown in Fig. 3.

It can be seen from Fig.~\eqref{fig:cd-steps} that using a great number of coordinate descent steps can achieve lower objective function value, however the computational time would increase. Therefore we should choose a suitable number of coordinate descent steps. In practice, we choose 3 steps of coordinate descent, which performs quit well in all experiments. Also, we can see from Fig.~\eqref{fig:cd-steps}(a) that SCC converges under coordinate descent steps.

\subsubsection{Learning Rates}
\begin{figure}[h]\centering
\subfigure[Stages 9-10]{\includegraphics[width=0.45\linewidth]{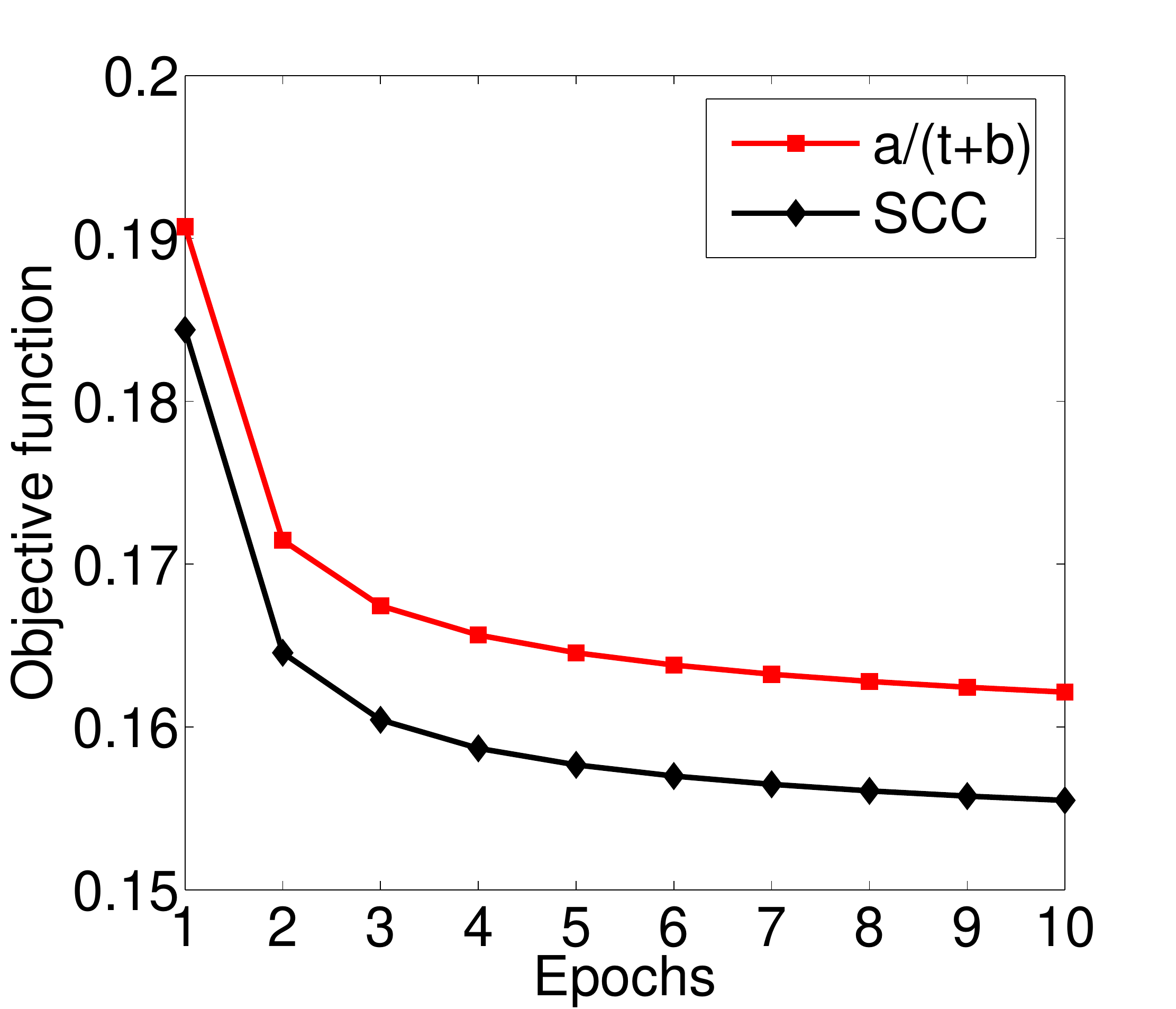}}
\subfigure[Stages 11-12]{\includegraphics[width=0.45\linewidth]{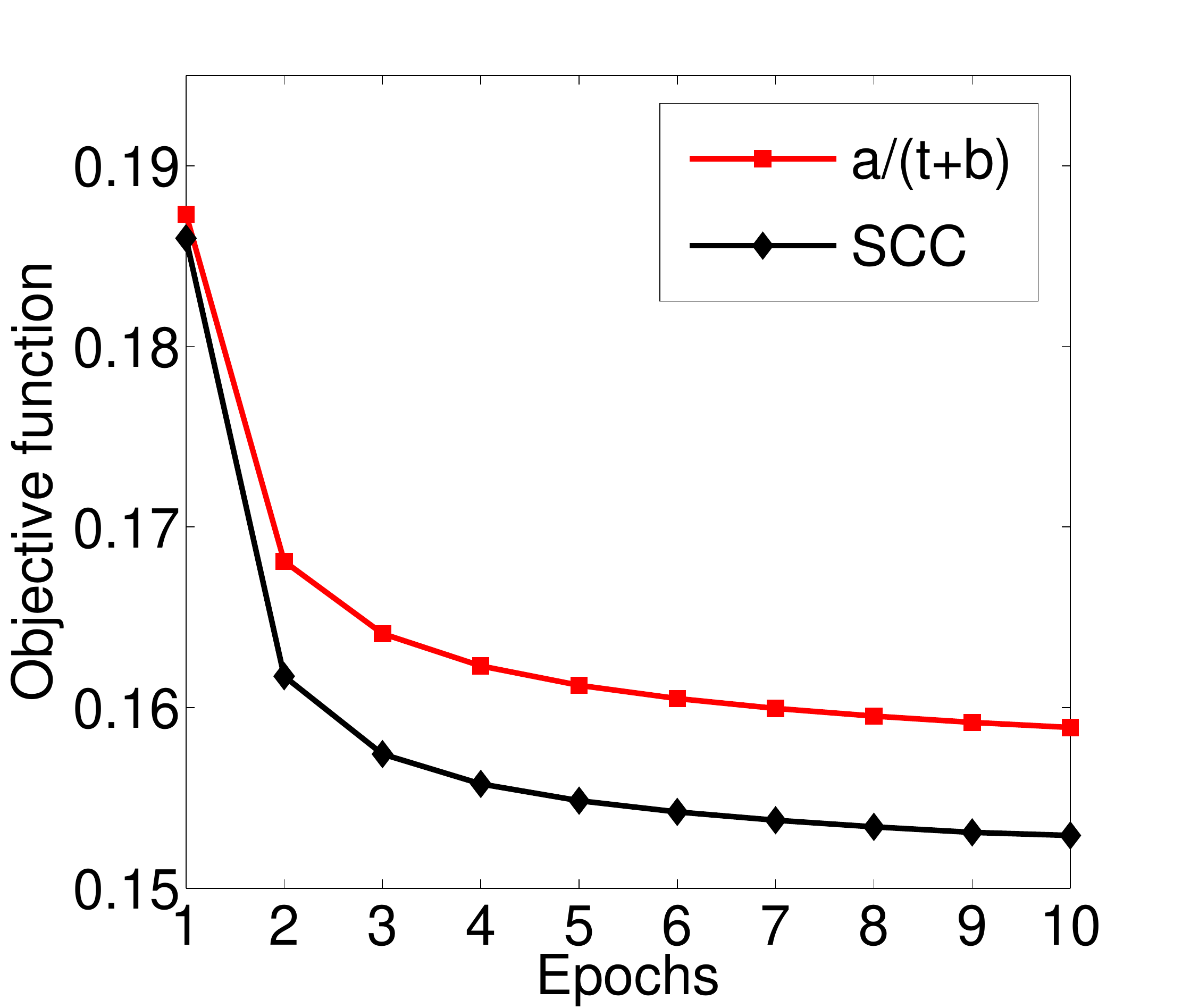}}
\caption{A comparison of different learning rates on two different stage ranges. From the figure we can see that the adaptive learning rate consistently performs better than the natural learning rates. Moreover, the adaptive learning rate converges faster than the natural learning rate.}\label{fig:learning-rates}
\end{figure}

We also studied the influence of the learning rate. We compared our adaptive learning rate with the natural learning rate on two different stage ranges. The natural learning rate is given as
$
\eta_t = \frac{a}{t+b},
$
where $a$ and $b$ are predefined parameters. We exhaustively search for the best parameter settings among $a \in [10^{-3}, 10^3]$ and $b \in [10^{-3}, 10^3]$ as determined by their lowest objective function value. We present the optimal result of the natural learning rate. It can be seen from the Fig.~\ref{fig:learning-rates} that the adaptive learning rate consistently performs better than the natural learning rate.

\subsection{Computational Time Comparison}

\begin{table*}[ht!]\caption{A comparison of SCC and OL on objective function values and computational time.}\label{tab:com-obj-time}
\centering
\vskip 0.15in
\begin{tabular}{|c|c|c|c|c|c|c| }
\hline  $128\times 500~$& Stages  & 4-6 & 7-8 & 9-10 & 11-12 & 13-17 \\
\hline
Objective function & OL & 0.1374 & 0.1495 & 0.1470 & 0.1465 & 0.1489\\ \cline{2-7}
value  & SCC & 0.1384 & 0.1503 & 0.1478 & 0.1474 & 0.1498\\
 \hline
Running time & OL & 15.71 & 7.81 & 8.60 & 30.64 & 31.14\\ \cline{2-7}
  (in hours)& SCC & {0.1439} & {0.0680} & {0.0748} & {0.2573} & {0.2616}\\
\hline
\end{tabular}
\vspace{-0.1in}

\centering
\vskip 0.15in
\begin{tabular}{|c|c|c|c|c|c|c| }
\hline  $128\times 1000$& Stages  & 4-6 & 7-8 & 9-10 & 11-12 & 13-17 \\
\hline
Objective function  & OL & 0.1317 & 0.1420 & 0.1398 & 0.1399 & 0.1424\\ \cline{2-7}
value & SCC & 0.1325 & 0.1429 & 0.1407 & 0.1408 & 0.1433\\
 \hline
Running time   & OL & 59.61 & 28.31 & 30.77 & 107.80 & 111.01\\ \cline{2-7}
 (in hours)& SCC & {0.1889} & {0.0900} & {0.0975} & {0.3399} & {0.3469}\\
\hline
\end{tabular}
\vspace{-0.1in}

\centering
\vskip 0.15in
\begin{tabular}{|c|c|c|c|c|c|c| }
\hline  $128\times 2000$& Stages  & 4-6 & 7-8 & 9-10 & 11-12 & 13-17 \\
\hline
Objective function & OL & 0.1266& 0.1358  & 0.1338 & 0.1342 &0.1365 \\ \cline{2-7}
 value& SCC &0.1278 &0.1366 &0.1349 &0.1353&0.1378\\
 \hline
Running time  & OL & 219.21 &102.64& 113.09&390.88& 397.34\\ \cline{2-7}
  (in hours)& SCC & {0.3666} & {0.1648} & {0.1902} & {0.6397} & {0.6401}\\
\hline
\end{tabular}
\vspace{-0.1in}
\end{table*}
\begin{figure*}[t]
\begin{center}
\subfigure[OL ($m = 500$)]{\includegraphics[width=.45\textwidth]{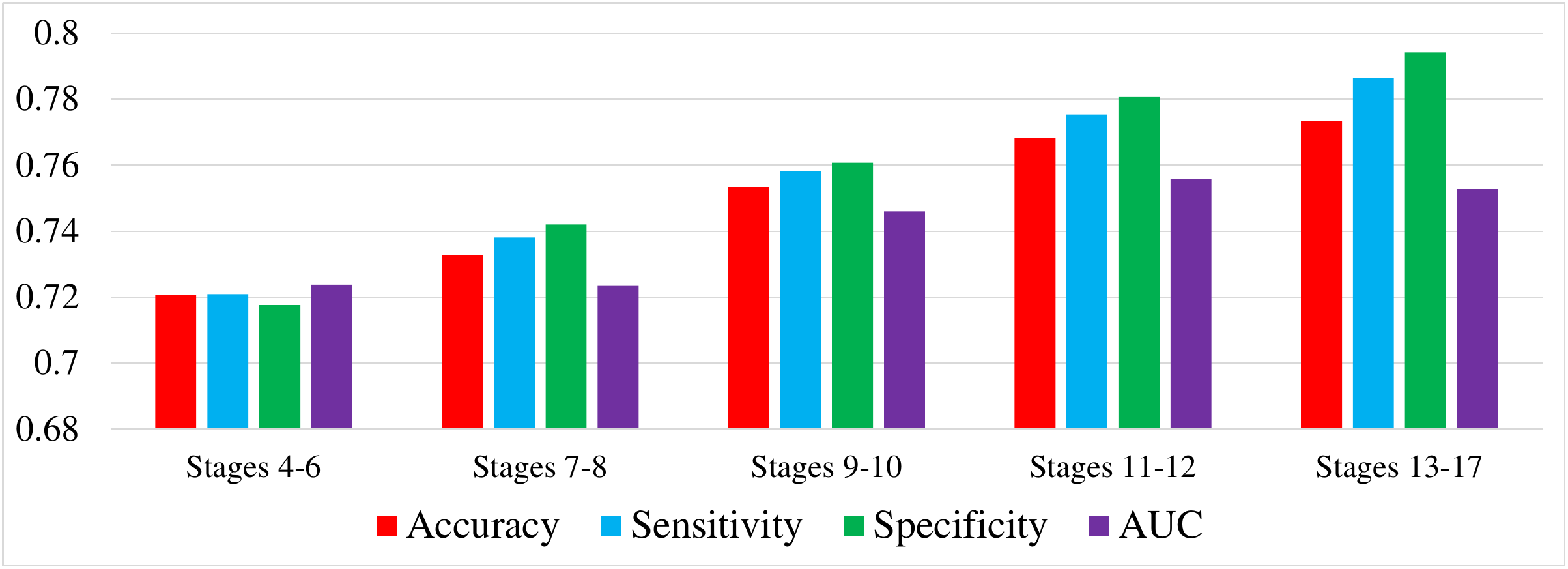}}
\subfigure[SCC ($m = 500$)]{\includegraphics[width=.45\textwidth]{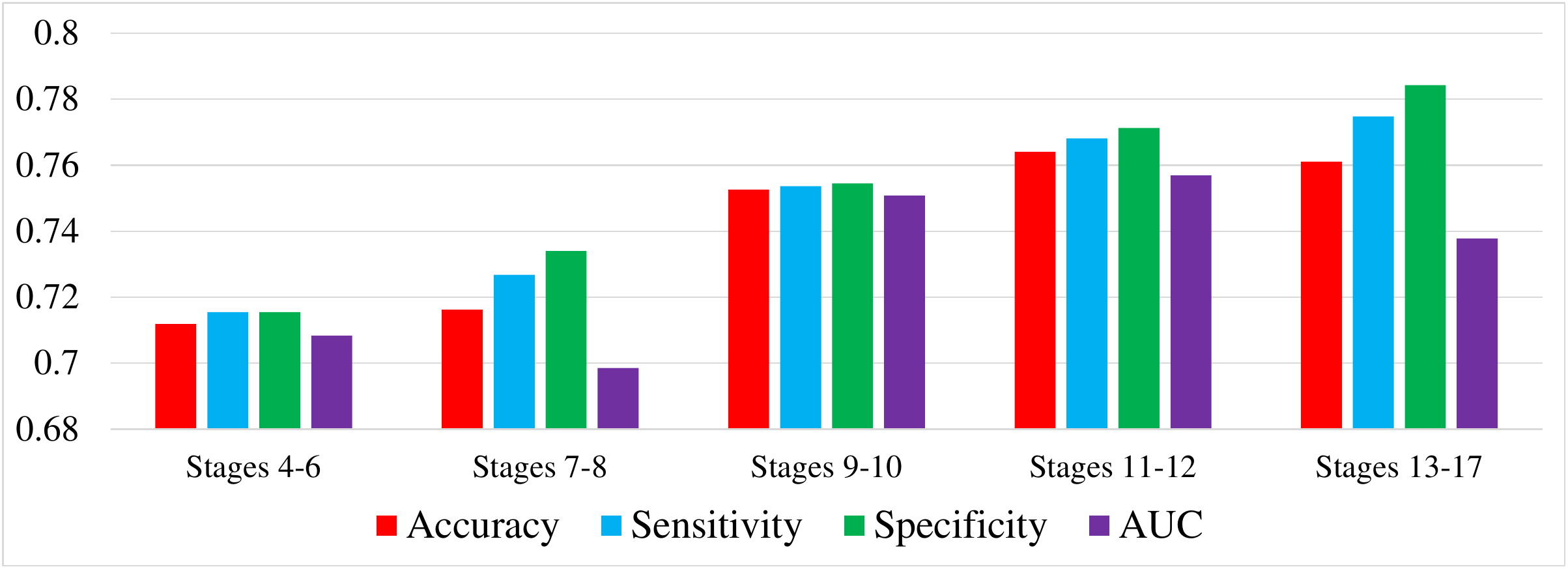}}\\
\subfigure[OL ($m = 1000$)]{\includegraphics[width=.450\textwidth]{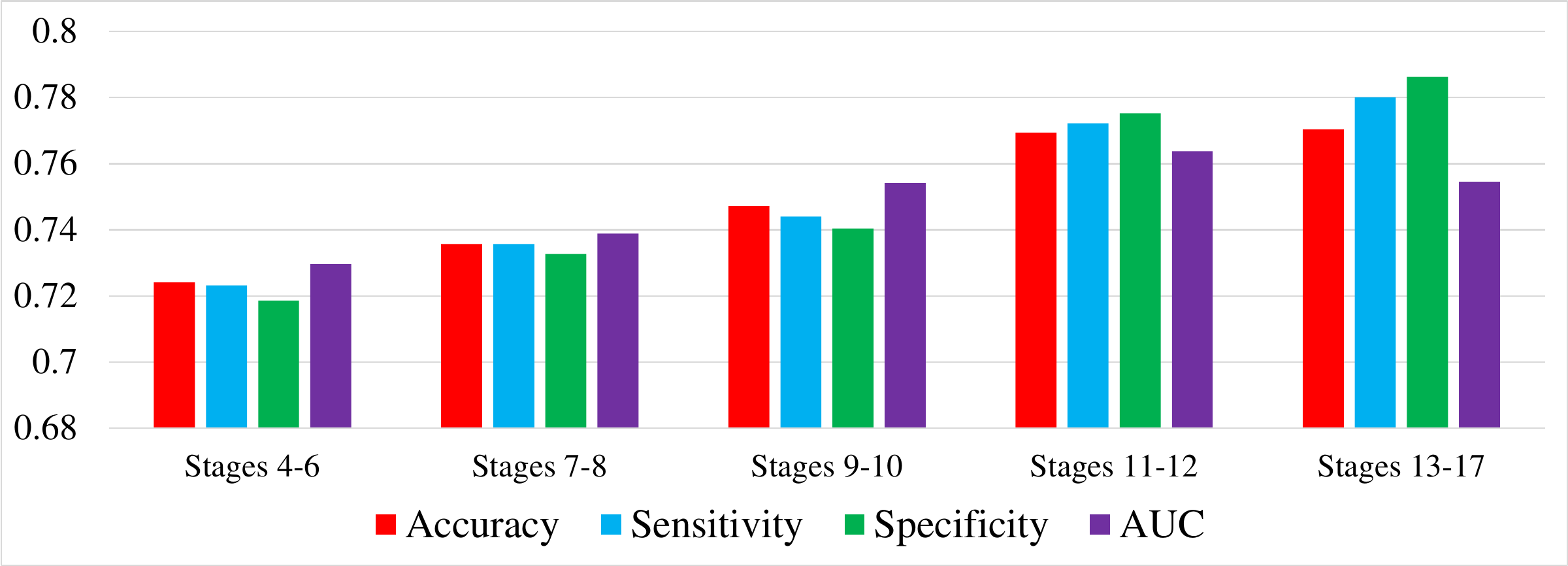}}
\subfigure[SCC ($m = 1000$)]{\includegraphics[width=.450\textwidth]{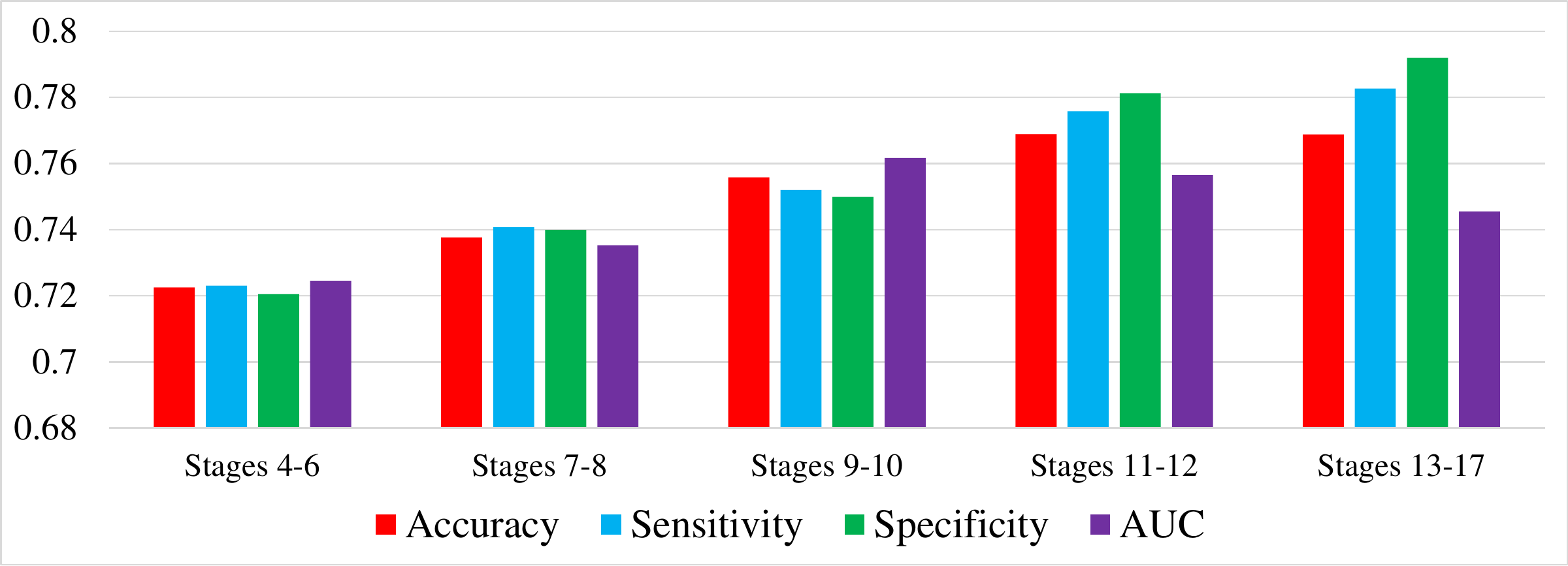}}\\
\subfigure[OL ($m = 2000$)]{\includegraphics[width=.450\textwidth]{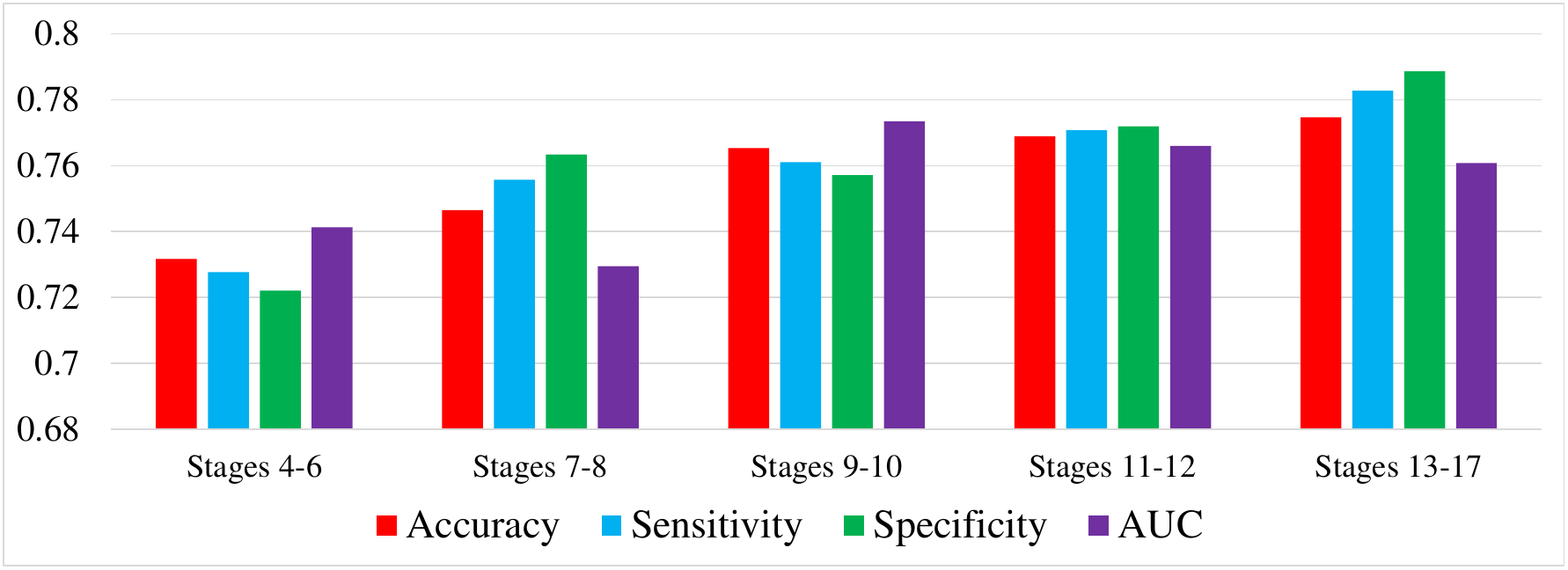}}
\subfigure[SCC ($m = 2000$)]{\includegraphics[width=.450\textwidth]{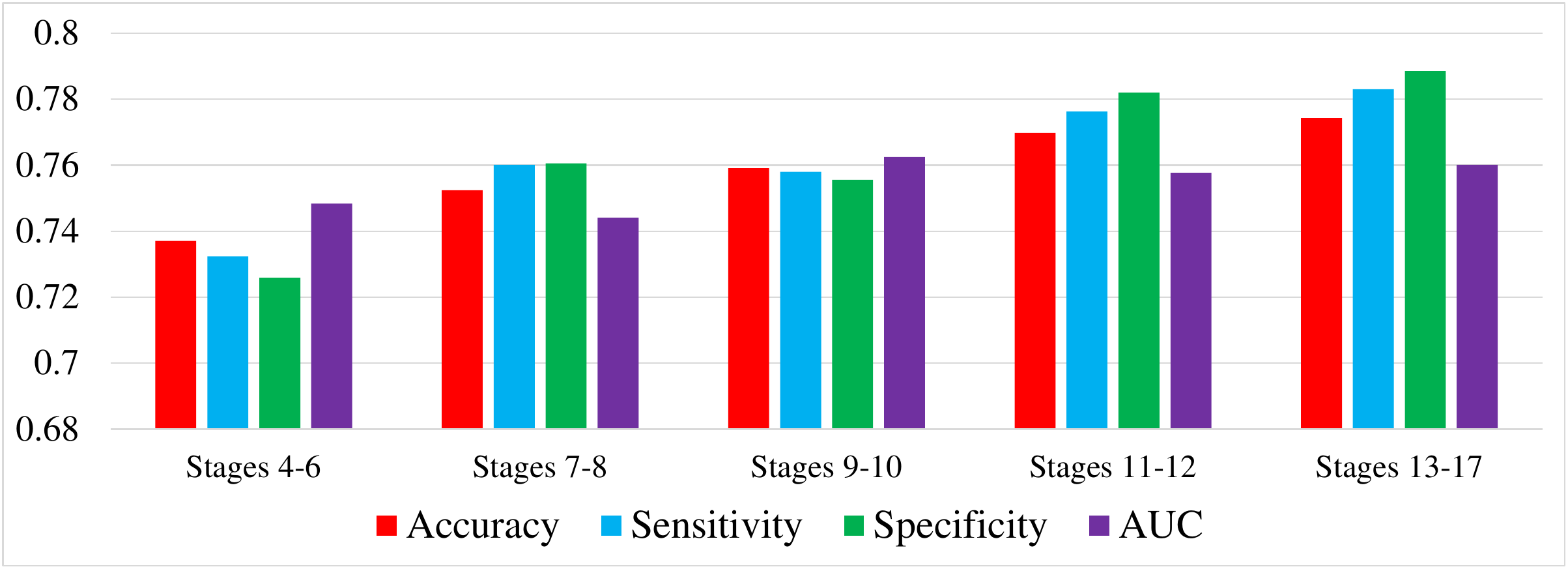}}
\vspace{-0.1in}
\caption{A comparison of SCC and OL on image annotation. (a)-(b), (c)-(d) and (e)-(f) show the image annotation results by using different dictionary sizes $500\times 128$, $1000\times 128$ and $2000\times 128$. It can be seen from the figure that when the dictionary size is $500 \times 128$, OL performs slightly better SCC. When the dictionary size is $1000\times 128$ or $2000 \times 128$, SCC and OL achieve comparable results.} \label{fig:image-annotation}
\vspace{-0.05in}
\end{center}
\end{figure*}
We first show the computational time of updating the dictionary and updating the sparse code. Table \ref{tab:com-z-D} shows the computational time of these two steps on three different dictionary sizes, i.e., $500\times 128$, $1000\times 128$ and $2000\times 128$. It can be seen from the table that SCC significantly reduces the computational time. Note that when the size of the dictionary increases, the computational time of OL increases rapidly. However, for SCC the computational time increases much slower compared to OL, especially the computational time of updating the dictionary. Therefore, SCC has a better scalability when dealing with large size dictionaries.

A computational time as well as an objective function value comparison is given in Table~\ref{tab:com-obj-time}. It can be seen from the table that SCC archives a very low objective function value, which is comparable with OL. Meanwhile, the computational time of SCC is much less than OL. Note that when the dictionary size increases, the objective function value decreases.

In this work we focus on the the single batch size setting, that is, we process one image patch in each iteration. We also compare our proposed SCC (with a batch size of 1) with mini-batch OL (with a batch size of 512). Our empirical results show that the mini-batch OL is about 3-4 times faster than SCC. Note that the mini-batch algorithms are usually faster than incremental algorithms. In addition, the implementation of mini-batch OL (SPAM\footnote{We use the code from the authors downloaded from the SPAM package: http://spams-devel.gforge.inria.fr/.}) is well optimized, making a direct time comparison difficult. We plan to develop the mini-batch extension of SCC and further optimize the code to improve its efficiency.

\begin{table}[ht!]\caption{A computational time comparison (in hours) of SCC and OL for different dictionary sizes.}\label{tab:com-z-D}
\centering
\vskip 0.15in
\begin{tabular}{m{2cm}<{\centering}ccccccc }
\hline
  &  \multicolumn{3}{c}{OL}  &\multicolumn{3}{c}{SCC}  \\ \cline{2-7}
Dictionary Sizes & 500  & 1000 & 2000 &  500 & 1000  & 2000\\
\hline
Update $\bZ$ & 2.58  & 8.83 & 38.75 &  0.22 & 0.39  & 0.71 \\
\hline
Update $\bD$ & 5.23  & 19.48 & 63.89 &  0.03 & 0.03  & 0.04 \\
\hline
Total & 7.81  & 28.31 & 102.64 &  0.25 & 0.42  & 0.75 \\
\hline
\end{tabular}
\vspace{-0.1in}
\end{table}

\subsection{Annotation Performance Comparison}
In this experiment, we compare the results of Drosophila gene image annotation by using learned features from SCC and OL. We tested all 5 stage ranges with different dictionary sizes, i.e., $500 \times 128$, $1000 \times 128$ and $2000 \times 128$. We choose four measurements: accuracy, AUC, sensitivity and specificity to evaluate the performance of different approaches. Comparison results for all 5 stage ranges by a weighted average of the top 10 terms are shown in Fig.~\ref{fig:image-annotation}.

It can be seen from the figure that when the dictionary size is $500 \times 128$, OL performs slightly better SCC. When the dictionary size is $1000\times 128$ or $2000 \times 128$, SCC and OL achieve comparable results. However, SCC is significantly faster than OL in this case. It might be worth noting that when the dictionary size increases, SCC and OL both improve their performance.

\section{Conclusions}
In this paper, we propose a new algorithm called Stochastic Coordinate Coding (SCC) to solve the sparse coding problem. In SCC, we perform a few steps of coordinate descent to update the sparse codes and use second order stochastic gradient descent to update the dictionary. The computational cost is further reduced by only updating the support of the sparse codes and the dictionary. Extensive experiments on Drosophila gene expression data sets have demonstrated the efficiency of the proposed algorithm. Compared to the state-of-art sparse coding algorithms, the proposed algorithm achieves one or two orders of magnitude speed-up~(see Table \ref{tab:com-obj-time}) when varying the dictionary size. The idea of combining coordinate descent and stochastic gradient descent can be applied to other problem settings. For example, we can extend the algorithm to solve the supervised sparse coding problem~\cite{NIPS2008_0775}. In addition, we plan to extend the algorithm from the single task learning setting to the multiple task learning setting. We also plan to develop the mini-batch implementation of SCC.


%
\bibliographystyle{abbrv}
\bibliography{Reference_dcai}  
%
%
\appendix
\section{Coordinate Descent for solving the lasso problem}

Given a data point $\bx = (x_1, \ldots, x_p)^T \in \mathbb{R}^p$ and a dictionary $\bD \in \mathbb{R}^{m\times p}$, the lasso problem is given as follows:
\begin{equation}\label{eq:lasso}
\min_{\bz} f(\bz) = \frac{1}{2}\| \bD\bz - \bx \|_2^2 + \lambda \| \bz \|_1,
\end{equation}
where $\bz = (z_1, \ldots, z_m)^T \in \mathbb{R}^m$.


If we freeze all components of $\bz$ except the $j$-th column $z_j$ in Eq.~\eqref{eq:lasso}. Let $\bd_j$ denote the $j$-th column of $\bD$, $d_{ij}$ the element of $\bD$ in the $i$-th row and $j$th column. We have
\begin{eqnarray*}
&&\argmin_{z_j} \frac{1}{2} \sum_{i=1}^p ( \sum_{j=1}^m d_{ij} z_j - x_i )^2 + \lambda \sum_{j=1}^m |z_j|\\
= && \argmin_{z_j} \frac{1}{2} ( z_j^2 - 2 b_j z_j + \|\bx\|_2^2 ) + \lambda |z_j|\\
= && \argmin_{z_j} \frac{1}{2} (   z_j - b_j)^2 + \lambda |  z_j |,
\end{eqnarray*}
where
$
b_j = \sum_{i=1}^p d_{ij} (x_i - \sum_{k\neq j} d_{ik} z_k)
$
and we have used the condition that each column of $\bD$ is unit norm.
Then $z_j$ has an exciplit optimal solution:
$
z_j =  h_{\lambda} (b_j),
$
where $b_j = \sum_{i=1}^p d_{ij} (x_i - \sum_{k\neq j} d_{ik} z_k)$ and $h$ is a soft thresholding shrinkage function or called the proximal operator of the $l_1$ norm \cite{Combettes2005-MMS}. It is defined as
\[
h_{\lambda} (v) = \left\{
\begin{aligned}
v + \lambda, &\quad v < -\lambda \\
0, & \quad-\lambda \leq v \leq \lambda \\
v - \lambda, &\quad\lambda < v
\end{aligned}
\right.
\]

Note that $
b_j = \sum_{i=1}^p d_{ij} (x_i - \sum_{k\neq j} d_{ik} z_k)
= \bd_j^T \bx - \bd_{j}^T \bD \bz + (\bd_j^T \bd_j) z_j
=\bd_j^T (\bx - \bD\bz)+ z_j.
$ Therefore, the computational cost of updating the $j$-th coordinate $z_j$ depends on computing the vector $\br = \bx - \bD\bz$ and the inner product $\bd_j^T \br$.





\end{document}